\documentclass[10pt,twocolumn,a4paper]{article}

\usepackage{cvpr}
\usepackage{mathptmx}
\usepackage{graphicx}
\usepackage{float}
\usepackage[caption = false]{subfig}
\usepackage{svg}
\usepackage{amsmath}
\usepackage{amssymb}

\usepackage{siunitx}
\usepackage{booktabs}
\usepackage[disable]{todonotes}

\usepackage{tikz}
\usetikzlibrary{positioning, matrix,calc,fit,decorations.pathmorphing,arrows,shapes}
\usepackage[utf8]{inputenc}

\newcommand{\autoe}[1]{\mathbb{A}_{#1}}
\newcommand{\clf}[1]{#1}
\newcommand{\nn}[2]{\autoe{#1}\circ\clf{#2}}

\usepackage[breaklinks=true,bookmarks=false]{hyperref}

\cvprfinalcopy 

\def\cvprPaperID{1838} 

\begin{document}

\title{What do Deep Networks Like to See?}

\author{Sebastian Palacio\thanks{Authors contributed equally} \quad Joachim Folz\footnotemark[1] \quad Jörn Hees \quad Federico Raue \quad Damian Borth \quad Andreas Dengel \\
German Research Center for Artificial Intelligence (DFKI)\\
TU Kaiserslautern \\
\texttt{first.last@dfki.de}}

\maketitle

\begin{abstract}
	We propose a novel way to measure and understand convolutional neural networks by quantifying the amount of input signal they let in.
	To do this, an autoencoder (AE) was fine-tuned on gradients from a pre-trained classifier with fixed parameters.
	We compared the reconstructed samples from AEs that were fine-tuned on a set of image classifiers (AlexNet, VGG16, ResNet-50, and Inception~v3) and found substantial differences.
	The AE learns which aspects of the input space to preserve and which ones to ignore, based on the information encoded in the backpropagated gradients.
	Measuring the changes in accuracy when the signal of one classifier is used by a second one, a relation of total order emerges.
	This order depends directly on each classifier's input signal but it does not correlate with classification accuracy or network size.
	Further evidence of this phenomenon is provided by measuring the \textit{normalized mutual information} between original images and auto-encoded reconstructions from different fine-tuned AEs.
	These findings break new ground in the area of neural network understanding, opening a new way to reason, debug, and interpret their results.
	We present four concrete examples in the literature where observations can now be explained in terms of the input signal that a model uses.
\end{abstract}

\section{Introduction}
\label{sec:intro}

Diagnostics for Deep Neural Networks often rely on measurements taken at the end of the processing pipeline.
Pinpointing issues with a network's architecture, learning process, and capacity typically depends on metrics based on the evolution of the loss function or on performance measurements like \textit{top-k accuracy}.
For example, to establish the presence of overfitting in a network, divergence of training and validation losses is considered as a good indicator.
Under these circumstances, there are general guidelines to follow like early stopping, loss regularization, acquiring more data or reducing the number of parameters in the model~\cite{bengio2012practical}.
Although these strategies are indeed effective against variance, they do not provide detailed insights on why the network failed to generalize in the first place.
To further understand the transformation of input samples into predictions, strategies have been proposed to look at the internal signals of networks~\cite{zeiler2014visualizing, simonyan2013deep}.
These provide insightful properties, but they serve a descriptive purpose rather than a predictive one.

\begin{figure}
	\footnotesize
	\includesvg[svgpath=img/,width=\linewidth]{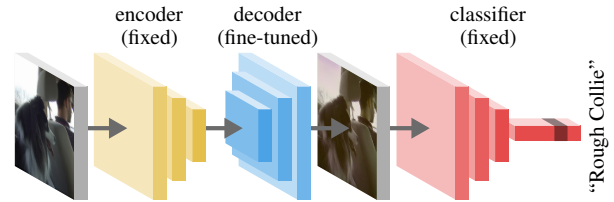}
	\caption{Overview of the proposed method and model. A pre-trained AE is fine-tuned with gradients flowing through a pre-trained image classifier whose parameters are fixed. After fine-tuning the combined network, images reconstructed by the AE preserve more information required by the classifier.}
	\label{fig:overview}
\end{figure}

Recently, Zhang \etal~\cite{zhang2016understanding} pointed out that our current understanding of how networks learn general features is incomplete and requires a different angle.
We propose a way to study neural networks based on quantifying the information contained in input samples that classification networks rely on.
In general, a deep convolutional neural network (DCNN) can be modeled as a function $y = f(x, \theta)$ where both the properties of the output $y$ and parameters $\theta$ have been widely studied.
In contrast, we introduce a novel way to examine the properties of the input $x$ and, more precisely, the amount of signal that the network itself takes in from $x$.

Intuitively a convolution operation with kernel size $k$ and striding $s$ covers the entire input spatially as long as $s \le k$.
In other words, such operations incorporate the complete input space.
In this work, we investigate how influential the input space is, not from the perspective of individual samples but rather of the model as a whole.
We found that, in practice, information from input samples that image classifiers use for prediction differs greatly between architectures.
These differences were measured by pre-training an AE on a large dataset and then fine-tuning its decoder using gradients from an image classifier with fixed parameters, as shown in Figure~\ref{fig:overview}.
Once the AE has been fine-tuned, predictions take place by passing samples through the AE first.
The effect of this specialized compression and reconstruction phase is two-fold:
First, information that is useful to the network gets preserved while irrelevant parts of the original input are canceled out.
Second, AEs fine-tuned in conjunction with classifiers learn to reconstruct the input in a way that attenuates any distracting aspects of the original input, such as noise.
Our approach offers the benefit that analysis happens in the input space and hence, any learned transformations by the AE are straightforward to visualize and interpret.
We have applied our method to five well-known DCNNs trained for image classification: LeNet5, AlexNet, VGG, Inception~v3, and ResNet-50.
Analyzing the input signal that these networks take in (characterized by the transformation of their corresponding fine-tuned AE), we found that they rely on different amounts of input signal.
Furthermore, this signal may be entirely different between classifiers.

Our main contributions are:
First, a model architecture and learning scheme that allows quantification of the input signal used by DCNNs.
Second, a relation of order that exists between high performance DCNNs concerning the input signal they use.
Third, we present an extensive, comparative evaluation of the input signal used by multiple state-of-the-art DCNNs backed up by well-established measures of information theory such as \textit{Mutual Information} (MI)~\cite{strehl2002cluster}.

\section{Related Work}

The need for interpretability in AI is growing with new laws and regulations~\cite{GoodmanRegulations} being introduced that govern its application. Given the ever growing body of evidence in favor of the effectiveness of Deep Networks, there is a pressing need for increased understanding how they work.
One of the first insights about their properties was that their features were general enough to perform well in different classification tasks~\cite{yosinski2014transferable, sharif2014cnn}.
Not only are these features transferable between tasks but they can also be distilled from an ensemble to a single model~\cite{hinton2015distilling}.

A second family of strategies has focused on understanding intermediate elements within networks like activation maps or convolutional filters.
Valuable insights came from visualizing said elements~\cite{zeiler2014visualizing}, but also from inspecting the degree of correlation that network filters share~\cite{li2015convergent}.

This kind of intermediate analysis has been extended all the way down to the input domain (otherwise known as \emph{activation maximization}).
Results show that not only are intermediate features encoding enough information to reconstruct the original input~\cite{mahendran2015understanding, dosovitskiy2016inverting}, but that it is also possible to identify areas within the input responsible for high prediction probabilities.
These areas can be modeled as a generic subset of the input~\cite{zhou2016learning, nguyen2016synthesizing} or as a collection of higher-level features~\cite{long2014convnets, netdissect2017}.
Moreover, this idea can be refined by explicitly accounting for background and foreground areas~\cite{simonyan2013deep} and even individual pixels~\cite{montavon2017explaining}.
Interestingly, studying the influence of individual pixels on deep image classifiers, gave rise to the research area of adversarial examples~\cite{szegedy2013intriguing}.

The use of AEs to remap the input space into better suited latent-space has been explored with general-purpose architectures like the variational auto-encoder~\cite{kingma2013auto}, as well as in specific tasks like noise suppression~\cite{vincent2010stacked}.
Moreover, it has been shown that initializing the weights of a network based on an auto-encoding scheme provides a good starting point for learning another task~\cite{masci2011stacked}.

Despite all these advances in understanding neural networks, recent work shows that we are still far from having a comprehensive notion about the learned features~\cite{zhang2016understanding}.
This highlights the need for new ways to analyze the capacity of a neural network (\eg, in terms of activation patterns or trajectory length~\cite{raghu2016expressive}).

\section{Methods}
This section describes the selection of the AE architecture and training scheme, and analysis performed on the resulting networks.
We pre-train a shared base AE and fine-tune it in conjunction with different classification networks to create tailored AEs.
We quantify and compare the information contained in images reconstructed by these fine-tuned AEs through
(1) relative changes of accuracy when input signals tailored to one classifier is used to measure the accuracy of another and (2) the amount of information that is present in images reconstructed by fine-tuned AEs.

\subsection{Autoencoder Selection}
\label{sec:segnet}
For our purposes we need an AE architecture that is capable of reliably capturing a large portion of information contained in input samples.
The reconstruction quality is primarily measured through the AE's loss.
Additionally, we define a validation measure that depends on the classification accuracy of a pre-trained network.
Informally, an AE produces good reconstructions if the accuracy of a pre-trained classifier does not change compared to the accuracy obtained by using the original inputs.
We chose SegNet~\cite{badrinarayanan2015segnet2} as the architecture for our AE, since it meets all the aforementioned requirements.
SegNet is a fully convolutional AE originally designed for semantic segmentation of RGB images.
It consists of two VGG-16~\cite{Simonyan14c} networks with batch normalization~\cite{ioffe2015batch} where the second half of the network has its layers reversed.
Max-pooling indices generated during the encoding stage are used to upsample activations in the decoder.
This enables the network to produce pixel-accurate reconstructions even though the smallest activation map is only $\frac{1}{32}$ of the input size.

\subsection{Autoencoder Pre-training}

\begin{figure}
	\includegraphics[width=\linewidth]{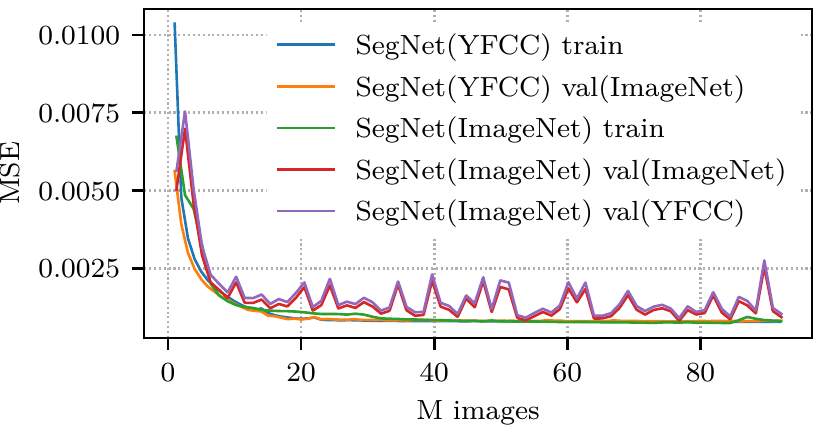}
	\caption{Mean-square reconstruction error of two SegNet networks trained on YFCC100m and ImageNet.}
	\label{fig:ae-training}
\end{figure}

As mentioned before, we first train the AE to minimize unsupervised reconstruction loss.
Ultimately, we do this so that transformed samples do not significantly change the accuracy of a pre-trained classifier (which will be addressed in Section~\ref{sec:normnet-training}).
Hence, we train the AE using the YFCC100m~\cite{YFCC100m}.
This dataset consists of $100$ million media objects taken from Flickr with roughly $99.2$ million images.
This set presents a comprehensive selection of the kinds of photos taken by a large group of people.
It constitutes the largest, publicly available image dataset to date.
We remove potential sources of noise in the form of placeholder images (\ie, samples that are listed in the dataset but were later removed by the user) and other non-photos (\eg, very small file size, single color) for a final count of $\approx 92.1$ million training images.
Using the YFCC100m as training set comes with some unique advantages.
On one hand, the concept of training step or epoch does not apply, since given the scale images do not need to be reused.
As shown in Figure~\ref{fig:ae-training}, the reconstruction error during training, smoothly converges to a good local minima before the dataset is used up.
On the other hand, learning without using any sample more than once makes the training loss an unbiased estimator of the model's ability to generalize.
A separate validation set is no longer required, since new samples have never been seen by the network.

We also compare this first run to the learning behavior of a second SegNet AE using ImageNet~\cite{ILSVRC15} instead of YFCC100m.
Both AEs were trained under similar conditions:
Input size of $256\times 256$ pixels, MSE loss, SGD with momentum $0.9$ and initial learning rate of $\eta = 0.01$.
When training on the YFCC100m, the reconstruction error is checked every 1 million images and, learning rate is reduced to $\eta' = 0.2\eta$ if the loss does not improve after two consecutive checks.
For comparability and fairness, when using ImageNet, we let the AE train for 72 epochs which amounts to $\approx 92.2$ million images that will be seen by the network during learning.
Although necessary for ImageNet but not for YFCC100m, we apply a common set of data augmentation operations (scale, rotation, shift, noise, blur, brightness, contrast, color, mirror) in both cases.
We also measured validation in two different ways:
For the AE trained on YFCC100m, we use ImageNet's validation set.
For the AE trained on ImageNet, we computed a validation scored based on ImageNet's validation set and, additionally, also on a subset of 50000 random samples taken from YFCC100m.
All training and validation curves are shown in Figure~\ref{fig:ae-training}.

At a glance we observe that both learning schemes reach good local minima for both training and validation.
However, using YFCC100m produces a faster-converging network which also reaches a consistently well-behaved lower bound for the loss function.
In contrast, training on ImageNet yields an asymptotic, unstable learning curve.
Validation sets closely followed the training curves with the AE trained on ImageNet and validated on a subset of YFCC100m producing the highest error.

These experiments indicate that using the YFCC100m as inital training for the AE is optimal, as it produces consistent, low error image reconstructions that generalize better than using a smaller dataset.
Having a well trained AE as a starting point for further training has been recommended~\cite{bengio2012practical, masci2011stacked} and proven to improve convergence.

\subsection{Autoencoder Fine-Tuning}
\label{sec:normnet-training}

Using the AE from Section~\ref{sec:segnet}, we now fine-tune it with gradients that originate from a pre-trained classification network with fixed parameters.
Intuitively, by letting the AE adapt the reconstruction function to produce samples that are likely to enhance classification performance, the AE will be rewarded for keeping the parts of the input that are relevant for inference.
Conversely, the fine-tuned AE will disregard any portions of the original input that do not have positive impact on the performance of the classifier.
An AE could then learn to represent all the information that is used by a classifier, if the accuracy of the latter does not decrease by using input reconstructions from the former.

Fine-tuning the AE occurs by feeding the reconstructed sample from the AE into a classifier, computing the prediction loss and, backpropagating the gradients all the way down to the input pixels and, further down the AE architecture.
Since the target of analysis is the classifier and not the combined AE-classifier network, the parameters of the classifier are not updated by the flowing gradients.
Only parameters belonging to the AE architecture get updated once gradients from the classifier start flowing into the AE itself.
Incoming gradients can be used to update parameters either on the encoding, the decoding, or both sides of the AE.
We found that the best results were obtained by updating only the weights belonging to the decoder.
Experiments that back up this argument are provided in the supplementary material.

To test the influence of architectural elements from the classification network in the fine-tuning process, we fine-tune copies of the pre-trained AE on four different classifiers pre-trained on ImageNet, as provided by the torchvision project\footnote{\url{https://github.com/pytorch/vision}, commit 10a387a}:
AlexNet~\cite{DBLP:journals/corr/Krizhevsky14},
VGG-16~\cite{Simonyan14c} with batch normalization,
Inception v3~\cite{DBLP:journals/corr/SzegedyVISW15},
and ResNet-50~\cite{DBLP:journals/corr/HeZRS15}.
We also added a version of LeNet-5~\cite{lecun1998gradient} modified to take input images of size $224\times 224$ pixels as a simple, lower bound in terms of classification accuracy.
These networks have been recognized as high performing models when they were first proposed.
Furthermore, they include a series of different structural elements that have played an important role to push the state of the art on image classification \eg, higher depth, batch normalization, inception modules, and residual connections.
We measure changes in accuracy when these networks use the unaltered images from the ImageNet validation set and when they use reconstructions from the AE of the same samples.
Moreover, we check if the reconstructed samples from the original pre-trained AE already encode all the information that each classifier uses for inference.

For the remainder of this paper, classifiers will be referenced by their first letter ($L$, $A$, $V$, $I$, and $R$) where applicable.
Furthermore, we will use the shorthand notation $\nn{i}{j}$ to indicate that the classifier $j$ is using input samples from an AE that has been fine-tuned with gradients provided by classifier $i$.
Additionally, $\autoe{S}$ will refer to the SegNet AE pre-trained on YFCC100m and, $\autoe{i}(x)$ will refer to a reconstruction of input sample $x$ using $\autoe{i}$.
We also define $\mathcal{C} = \{L,A,V,I,R\}$, the set of all classifiers evaluated.

\begin{table}
	\caption{Center-crop, single-scale accuracies on ImageNet validation set for: original classifier, classifier using reconstructions from $\autoe{S}$ and, classifier using reconstructions from the AE that was fine-tuned in conjunction with the classifier.}
	\label{tab:accuracy-classifiers}
	\hspace*{\fill}
\begin{tabular}{rrrrr}
	\toprule
	Network            & top-1 & diff      & top-5 & diff \\
	\midrule
	$L$            & $32.30$        &   & $54.63$ &          \\
	$\nn{S}{L}$     & $31.08$ & $-1.22$ & $53.01$ &$-1.62$ \\
	$\nn{L}{L}$     & $34.85$ & $+2.55$ & $57.79$ & $+3.61$ \\
	\midrule
	$A$            & $54.96$      &     & $77.98$   &        \\
	$\nn{S}{A}$     & $51.89$ & $-3.07$ & $75.52$ &$-2.46$ \\
	$\nn{A}{A}$     & $56.13$ & $+1.17$ & $78.96$ &$+0.98$ \\
	\midrule
	$V$          & $71.35$ & &$90.50$ &\\
	$\nn{S}{V}$   & $67.59$ &$-3.76$ & $87.95$ &$-2.55$ \\
	$\nn{V}{V}$   & $71.65$ &$+0.30$ & $90.55$& $+0.05$ \\
	\midrule
	$R$          & $74.02$ && $92.01$& \\
	$\nn{S}{R}$   & $71.19$ &$-2,83$ & $90.23$ &$-1.78$ \\
	$\nn{R}{R}$   & $74.94$ &$+0.92$ & $92.27$ &$+0.26$ \\
	\midrule
	$I$       & $77.12$ && $93.25$& \\
	$\nn{S}{I}$& $74.42$ & $-2.70$ & $91.87$ & $-1.38$ \\
	$\nn{I}{I}$& $76.71$ & $-0.41$ & $93.03$ & $-0.22$ \\
	\bottomrule
\end{tabular}
\hspace*{\fill}

\end{table}

Table~\ref{tab:accuracy-classifiers} shows top-1 and top-5 accuracies for the aforementioned classifiers before and after pre-pending them with their correspoinding fine-tuned AE.
As described above, we also computed the baseline $\nn{S}{i}, i\in \mathcal{C}$ for reference.
We see that the baseline lays between 1.2 and 3.7 points below the accuracy of the classifiers alone.
Although this is not a dramatic drop, it does indicate that some information has been lost during the reconstruction of input samples.
Taking into account that the difference between $x$ and $\autoe{S}(x)$ is quite small, we can infer that the missing signal that is relevant for inference has to be small as well.
Notwithstanding, fine-tuning $\autoe{S}$ on any classifier, already makes up for the initial loss of information and, for most cases, even surpasses the performance of the classifiers alone.

\subsubsection{Encoded Representations of Fine-Tuned AEs}
\label{sec:normnet-training-finetune}

\begin{figure*}
	\includegraphics[width=\textwidth]{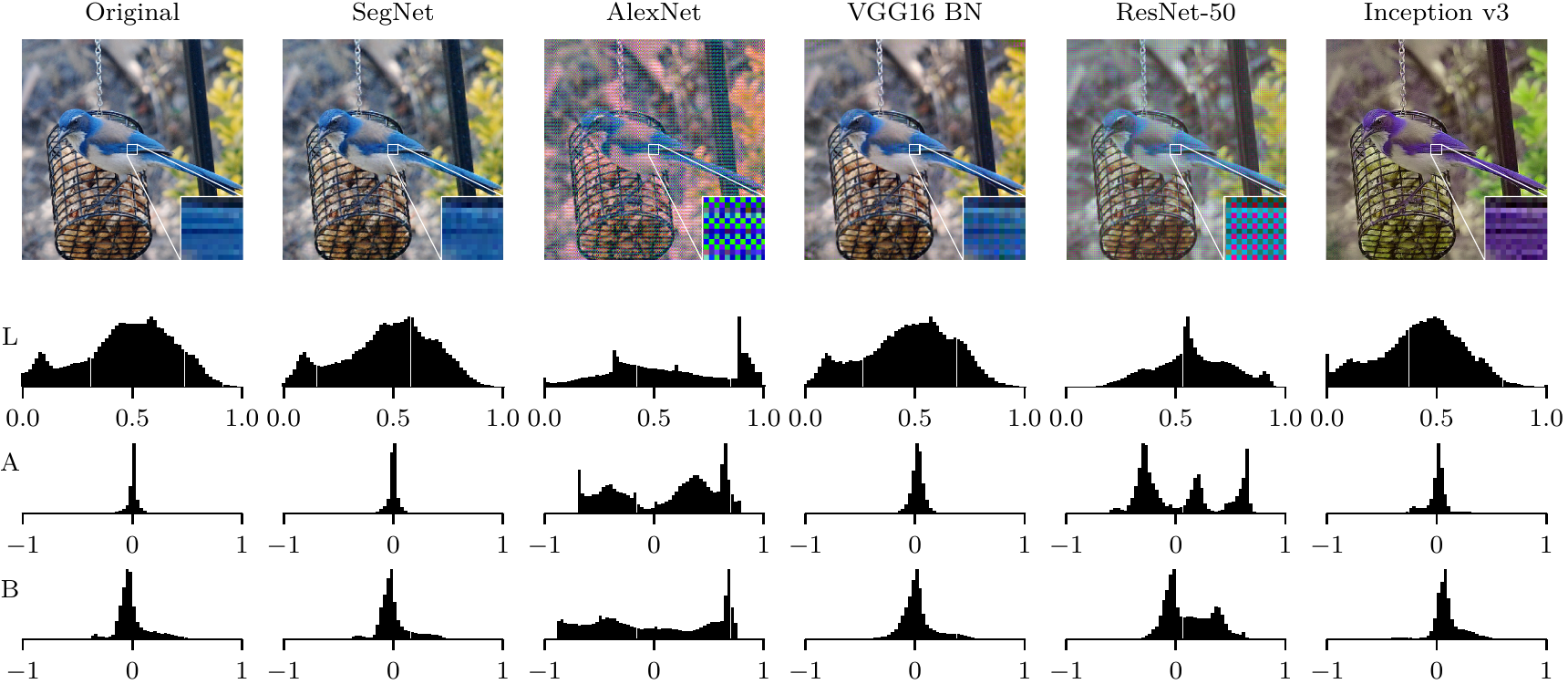}
	\caption{Example taken from ImageNet's validation set. Each image corresponds to a reconstruction produced by one of the fine-tuned AEs. Below each reconstruction, global histograms of the image in LAB space are shown. Zoomed portions of the image reconstructions are provided, showing the emergent patterns in more detail.}
	\label{fig:normnet-reconst-hist}
	\vspace{-0.2em}
\end{figure*}

Once we obtain all $\autoe{i}, i\in \mathcal{C} \cup \{S\}$ by fine-tuning, we can have a look at the reconstructed images.
Figure~\ref{fig:normnet-reconst-hist} shows an original image and the corresponding reconstructions by each $\autoe{i}$
(for more examples, please refer to the supplementary material).
Below each reconstructed image, we visualize channel-wise histograms in LAB color space.
These histograms allow us to identify global changes in perceived brightness and opposing colors.
The ``L'' channel represents the luminance, ``A'' encodes color changes between green and magenta and, ``B'' encodes changes between blue and yellow.
We chose this representation because it relates closely to how human vision works.
It neatly separates perceived brightness from color and does not suffer from range singularities like HSV.

The observed transformations are unique for each classifier and are easily identified as large changes in all three channels when compared to the original image.
Changes introduced by $\autoe{V}$ are the smallest among all AEs.
$\autoe{A}$ and $\autoe{R}$ introduce consistent checkerboard artifacts over the entire input space.
These artifacts appear as peaks in the histograms in channels ``A'' and ``B''.
$\autoe{R}$ and $\autoe{I}$ compress the range of the luminance, making dark areas brighter and bright areas darker.
$\autoe{I}$ introduces a strong shift in ``B'' values that manifest as a lack of blues and yellow/brown tint.
$\autoe{A}$ produces images with a distinct pink hue.

\subsubsection{Emergent Resilience to Noise}
\begin{figure}
	\centering
	\includegraphics[scale=1]{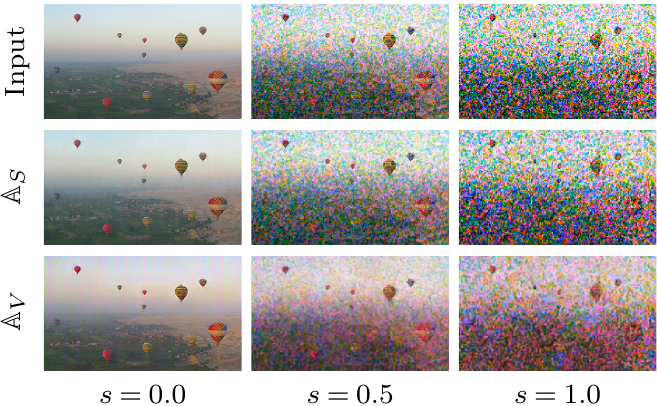}
	\caption{Example for additive noise used to evaluate performance in noisy environments for different values of the strength parameter $s$. Top is the original input image. Middle and bottom rows are reconstructions of $\autoe{S}$ and $\autoe{V}$.}
	\label{fig:noisy-images}
\end{figure}
\begin{figure}
	\includegraphics{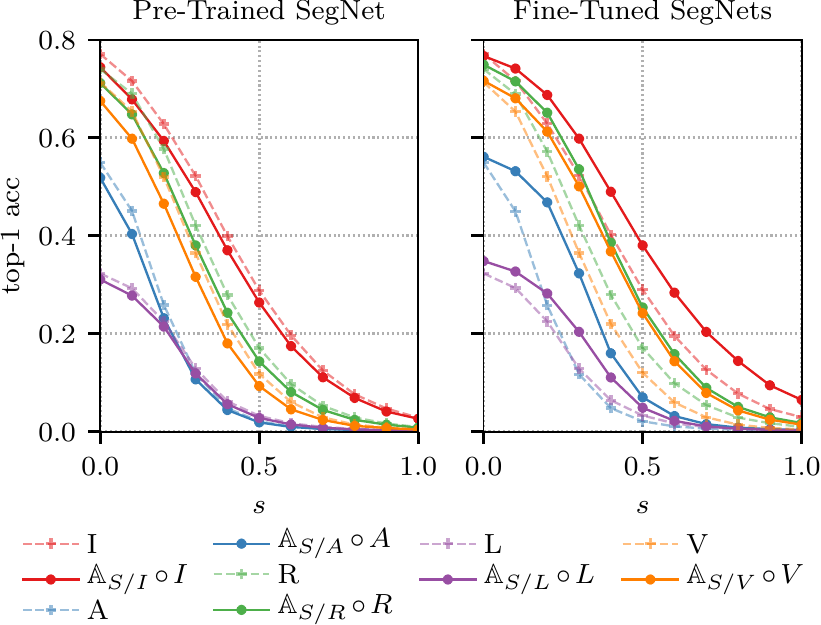}
	\caption{Top-1 accuracy on ImageNet's validation set under increasing noise strength $s$. Dashed lines correspond to networks evaluated on the original inputs. Solid lines depict evaluation of the networks using reconstucted inputs from fine-tuned AEs. Left: behavior with reconstructions from $\autoe{S}$. Right: behavior with reconstructions from $\autoe{i}$, $i \in \mathcal{C}$.}
	\label{fig:noisy-accuracy}
\end{figure}
We noticed that a consequence of fine-tuning AEs, as proposed in this work, is that their reconstructions are preserving more information under noise than the original inputs.
We tracked the top-1 accuracy of the classifiers in $\mathcal{C}$ on ImageNet's validation set under an increasing additive uniform noise drawn from $\mathbb{U}[-s, s]$ with strength $s\in [0, 1]$.
We compared the results to a similar setup where the accuracy was computed using reconstructed samples from $\autoe{i \in \mathcal{C} \cup \{S\}}$.

For the first experiment (dashed lines in Figure~\ref{fig:noisy-accuracy}), accuracy starts decays steadily, as soon as noise gets introduced.
Using reconstructions from $\autoe{S}$ (left of Figure~\ref{fig:noisy-accuracy}), worsens performance even more, suggesting that the AE itself has not learned a denoising transformation.
Note that these results are aligned with initial findings presented in Section~\ref{sec:normnet-training-finetune} where classification with reconstructed samples of $\autoe{S}$ alone were already lower than the correspoinding baseline.
In contrast, reconstructions from fine-tuned AEs are consistently more resilient to noise, even when compared to the stand-alone classifiers.
Accuracy for classifiers of the form $\nn{i}{i}$ present higher accuracies for all levels of noise, indicating that all $\autoe{i \in \mathcal{C}}$ have indeed learned a representation of the information that is useful to the classifier it was fine-tuned on.
The most pronounced difference can be seen with AlexNet, which falls below the accuracy of LeNet-5 at $s=0.3$, yet it manages to remain above LeNet-5 when using inputs from $\mathbb{A}_A$.
This measurable resiliency to noise is also visually perceptible as shown in Figure~\ref{fig:noisy-images}.

\subsection{Measuring Information through Classifiers}
\label{sec:crossvalidation}

We explored the relationship between different encodings defined by fine-tuned AEs.
By observing the wide visual differences between image reconstructions from all $\autoe{i \in \mathcal{C}}$, it is clear that each classifier prefers different reconstructions and thus, different information.
Hence, we measured changes in accuracy for each classifier when using input reconstructions from AEs that were fine-tuned on other classifiers.
More formally, we evaluate the accuracy of $\nn{i}{j}, \forall i,j \in \mathcal{C}$.
Results are summarized in Table~\ref{tab:cross-validation}.

\begin{table}
	\caption{Cross-validation accuracies on ImageNet validation set for classifiers that receive inputs from AEs fine-tuned on different models.}
	\label{tab:cross-validation}
	\hspace*{\fill}
	\begin{tabular}[]{lccccc}
		\toprule
		  & L & A & V & I & R \\
		\midrule
		$\mathbb{A}_L$ & 0.3484 & 0.3077 & 0.0416 & 0.4352 & 0.4730 \\
		$\mathbb{A}_A$ & 0.0211 & 0.5613 & 0.0097 & 0.5375 & 0.0925 \\
		$\mathbb{A}_V$ & 0.2929 & 0.5362 & 0.7163 & 0.7400 & 0.7300 \\
		$\mathbb{A}_I$ & 0.1829 & 0.3024 & 0.4555 & 0.7671 & 0.4540 \\
		$\mathbb{A}_R$ & 0.0163 & 0.4972 & 0.0710 & 0.7249 & 0.7494 \\
		\bottomrule
	\end{tabular}
	\hspace*{\fill}
\end{table}

\begin{figure}
	\includegraphics[width=\linewidth]{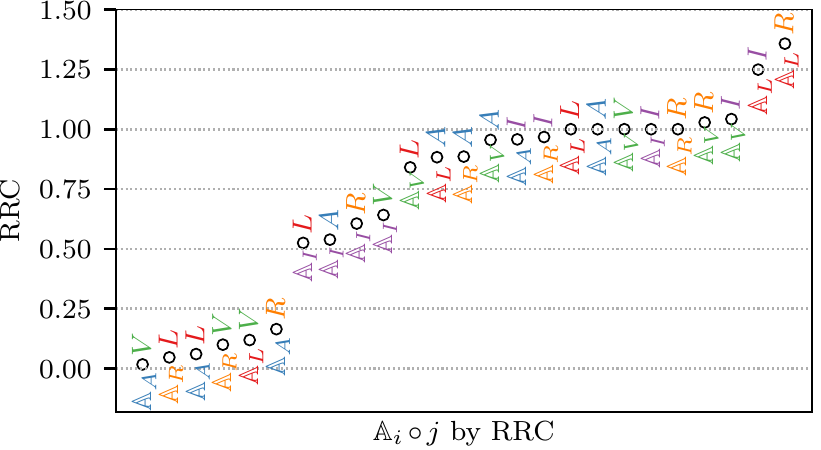}
	\caption{Every tested $\nn{i}{j}$ combination, sorted by RRC. Parts are color coded according to their associated classifier: L red, A blue, V green, I purple, R orange.}
	\label{fig:rrc-min}
\end{figure}

We observed that accuracy drops consistently for each classifier when they use other AEs.
However, some combinations of $\nn{i}{j}$ tend to preserve the accuracy of the lowest performing model in that combination \ie, $\exists_{i\neq j}: acc(\nn{i}{j}) \approx min(acc(i), acc(j))$.
Informally, this effect can be interpreted as networks that use \emph{at least the signal that the lowest performing model uses}.
To quantify this, we define the \emph{relative rate of change (RRC)} of an AE-classifier pair as follows:
$RRC(\nn{i}{j}) = \frac{acc(\nn{i}{j})}{m(i, j)}$ where $m(i, j) = min(acc(i),acc(j))$.
Computing RRC values on the cross-validation experiment reveals which combinations of AEs and classifiers preserve more signal, as shown in Figure~\ref{fig:rrc-min}.
From this curve, we see how the input signal used by VGG is enough to make both ResNet and Inception perform better than VGG itself.
Additionally, the signal from Inception~v3 seems to be quite different from the one used by any other model, as none of the models in $\mathcal{C}\setminus \{I\}$ performed well (\ie, below 0.65 of their original accuracy) using $\autoe{I}$.

\begin{figure}
	\footnotesize
	\hfill
	\subfloat[$t=0.1$]{\includesvg[svgpath=img/,width=0.141\linewidth]{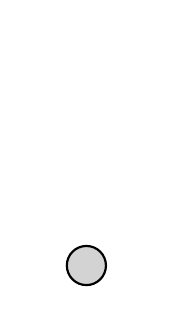}} 
	\hfill
	\subfloat[$t=0.2$]{\includesvg[svgpath=img/,width=0.129\linewidth]{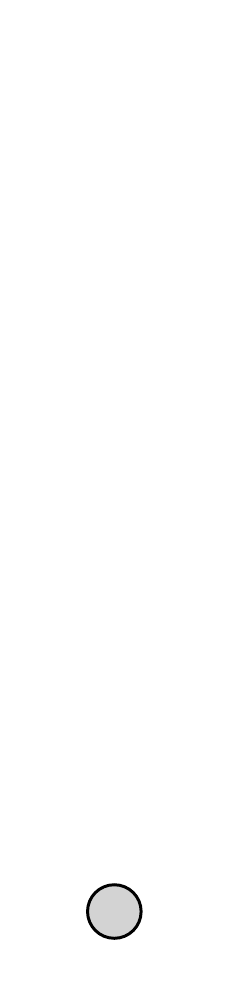}}
	\hfill
	\subfloat[$t=0.8$]{\includesvg[svgpath=img/,width=0.1315\linewidth]{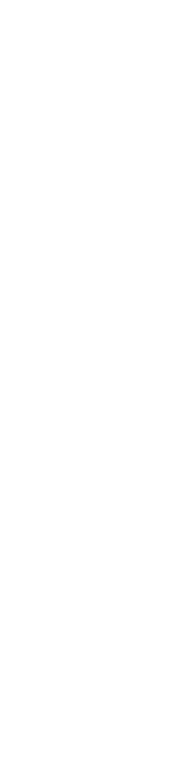}}
	\hfill
	\subfloat[$t=0.9$]{\includesvg[svgpath=img/,width=0.257\linewidth]{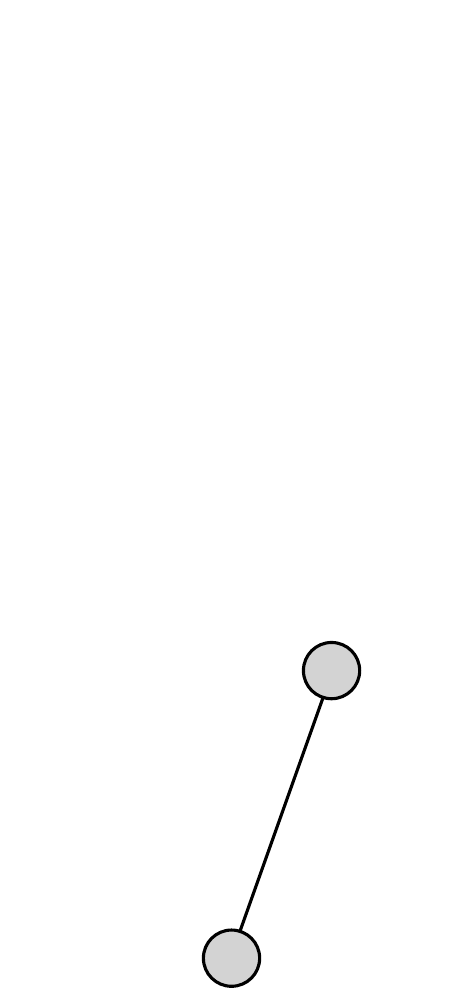}}
	\hfill
	\caption{FC lattices for different RRC thresholds. Attributes (signal) are always below the nodes, objects (classifiers) are on top.}
	\label{fig:lattices}
\end{figure}

To further examine the relation between AEs and classifiers, we use formal concept analysis (FCA) \cite{wille1982restructuring} to derive a hierarchical ontology between all possible combinations of them.
FCA is of particular interest here because it allows us to model partially order sets (under the inclusion operation $\subseteq$).
To this end, we need to define a \textit{formal context} $\mathbf{K} = (\mathbf{G}, \mathbf{M}, \mathbf{I})$ where $\mathbf{G}$ is a set of \textit{objects}, $\mathbf{M}$ is a set of \textit{attributes} and $\mathbf{I}$ is a binary relation between elements of $\mathbf{G}\times \mathbf{M}$ that expresses whether $\mathbf{G}$ has the attribute $\mathbf{M}$ or not.
Formal concepts of $\mathbf{K}$ are object-attribute subset pairs $(S_A, S_B)$ such that $S_A' = S_B$ and $S_B' = S_A$, where $S_A' = \{m \in \mathbf{M} | \forall_{g\in \mathbf{G}} g\mathbf{I}m \}$ and $S_B' = \{g \in \mathbf{G} | \forall_{m\in \mathbf{M}} g\mathbf{I}m\}$.
The lattice of formal concepts for $\mathbf{K}$ is constructed by ordering paris of formal concepts $(S_A, S_B), (S_C, S_D)$ under the operation $\le: (S_A, S_B)\le (S_C, S_D) \leftrightarrow S_A \subseteq S_C$.
Please refer to \cite{wolff1993first} for a more intuitive introduction on FCA.
Let $\mathbf{G} = \mathcal{C}$ and $M = \{\autoe{i \in \mathcal{C}}\}$.
Finally, let $\mathbf{I} = \{(i, j): RRC(\nn{i}{j}) \ge t\}$ for a given threshold $t$.
In other words, we convert the table of RRC values into a binary relationship between AEs and classifiers by applying a threshold to it.
We generate lattices for the FCs at thresholds $t\in\{0.1, 0.2, 0.8, 0.9\}$, shown in Figure~\ref{fig:lattices}.
For any two nodes connected by an edge, the upstream connection can be interpreted as \emph{the signal encoded by $\autoe{i}$ is used by classifier $j$}.
Looking at $t\in \{0.1, 0.9\}$ gives an idea of the most sensitive and robust changes in signal behavior since they are close to the upper and lower bound in the range of RRCs.
Similarly, FCs for $t\in \{0.2, 0.8\}$ characterize the largest changes in signal intake (\ie, they lay in between the largest gaps) among classifiers.
Note that any value of $t$ between those intervals ($[0.17 - 0.52]$ and $[0.66 - 0.84]$) yield the same FC, thus the same lattice.
This is important for establishing more precise lower and upper bounds between signals and classifiers later on.

Looking at the first lattice in Figure~\ref{fig:lattices}a, we can see that the signal from Resnet's AE is already not enough to make VGG or LeNet reach an accuracy that can go above 10\% of what either of them achieves on their own, using their corresponding AE.
This is especially surprising for VGG, considering how similar its performance is compared to ResNet.
Conversely, in Figure~\ref{fig:lattices}d we see how the signal of VGG is enough for making Resnet's performance at least 90\% of what VGG--the lowest performing of the two--is originally capable of.
For Inception, we see that it shares up to 50\% of the signal that the other classifiers are picking up on.
However, the remaining half of the signal used by all other networks appears to be completely different from the one Inception uses to achieve its high performance.
As a general trend, VGG stands out as the classifier making the most exhaustive use of the input signal.
The AE fine-tuned on VGG preserves a signal that makes all other classifiers keep a performance of 80\% and higher.

This analysis shows that all networks extract features based on a common portion of the input signal but said portion, can be as small as 10\%, as pointed out earlier for Resnet.
Note how any FC for thresholds between 0.2 and 0.8 yield lattices describing a totally ordered set.
The only changing element is the AE fine-tuned on Inception.
Such a total order exposes an unexplored aspect that networks are sensitive to, namely that DCNNs are only extracting features from a reduced portion of the input signal.
We interpret this hierarchy as the amount of general or specialized signal used by a network.

\subsection{Measuring Information through Image Reconstructions}

We validate the pattern found in the cross-validation experiments from Section~\ref{sec:crossvalidation} by measuring loss and preservation of information between input reconstructions.
We use the normalized mutual information (nMI)~\cite{strehl2002cluster} measure to calculate bounded values reflecting relative changes in the information that is preserved or lost when input samples are passed through each AE.
There are two complementary cases to be considered, as shown in Figure~\ref{fig:nmi-idea}:

\begin{figure}
	\centering
	\includegraphics[width=0.8\linewidth]{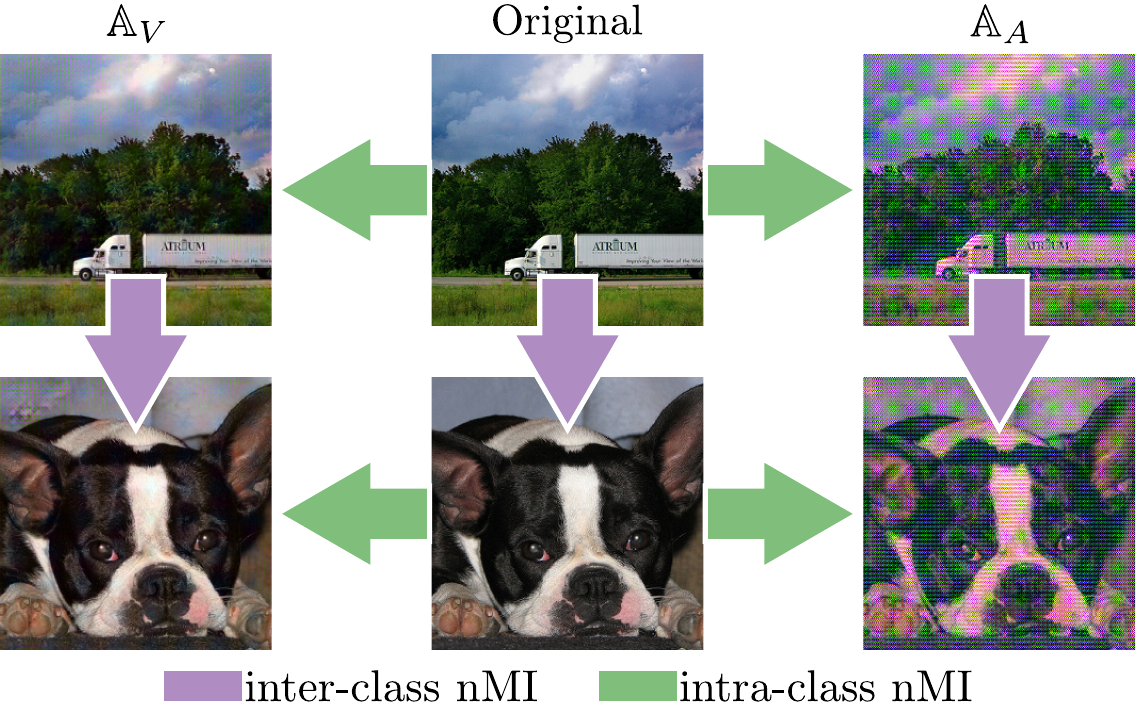}
	\caption{Difference between intra-class nMI and inter-class nMI. The former measures information between the original sample and reconstructions from different AEs. The latter measures reconstructions of different samples using the same AE.}
	\label{fig:nmi-idea}
\end{figure}

\vspace{-1.5em}
\paragraph{Intra-class nMI:} compare reconstructions from each fine-tuned AE to the original sample.
Reconstructed samples should preserve as much information as possible from the original image and hence, a high correspondence is expected.
To compute the intra-class nMI, an input sample is passed through all AEs and the nMI is computed with respect to the original sample.
In other words, $\texttt{intra-nMI}(x, \autoe{i}) = \texttt{nMI}(x, \autoe{i}(x))$.
For each AE, the average nMI and the standard deviation of all samples in the validation set of ImageNet (50000 samples) are reported.

\vspace{-1.5em}
\paragraph{Inter-class nMI:} compare reconstructions of two different samples using a single fine-tuned AE.
Reconstructed samples of two independent images should yield low nMI values.
Therefore, a low correspondence is expected.
To compute the inter-class nMI, two samples are drawn at random, passed through the same AE and, the nMI is calculated between those two reconstructions.
More formally, $\texttt{inter-nMI}(x_1, x_2, \autoe{i}) = \texttt{nMI}(\autoe{i}(x_1), \autoe{i}(x_2)$.
For each AE, the average nMI and the standard deviation of consecutive sample pairs over the entire set of Imagenet (25000 pairs) are reported.

\begin{figure}
	\includegraphics[width=\linewidth]{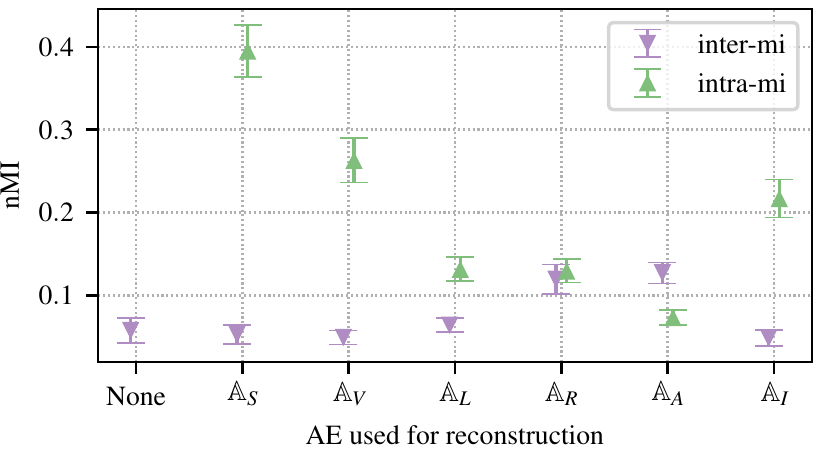}
	\caption{Normalized mutual information for input samples reconstructed from different AEs. \textit{Intra-class} nMI measures information between reconstructions of the same sample}
	\label{fig:nmi}
\end{figure}

Results shown in Figure~\ref{fig:nmi} that $\autoe{S}$ is able to preserve the highest amount of the input signal.
Moreover, there is a well-defined order with respect to classifiers whose fine-tuned AEs preserve more information.
Regardless of the variant of nMI used for comparison, all fine-tuned AEs can be sorted as follows: $\autoe{V} > \autoe{I} > \autoe{L} > \autoe{R} > \autoe{A}$.
Note that both $\autoe{A}$ and $\autoe{R}$ show comparatively high amounts of inter-class nMI, which is consistent with earlier observations of highly regular patterns in their reconstructions as described in Section~\ref{sec:normnet-training-finetune}.
The resulting totally ordered set aligns closely to the attributes in the lattice of Figure~\ref{fig:lattices}b which also describes a totally ordered set.

This analysis provides further evidence of the existence of a pattern, where different networks are using of more or less information from the original input.
Furthermore, considering that classification accuracy dropped consistently for all classifiers when the original AE was used, we can infer that the lost information, though irrelevant in most cases, is indeed used by all classifiers.
The nature of information reconstructed by $\autoe{S}$ and its fine-tuned versions suggests that classifiers are using parts of the inputs that are less relevant for accurate reconstruction.

\subsection{Understanding Previous Work based on Signal}
We have so far been able to quantify the amount of signal different classifiers use for their predictions.
Another way of interpreting our findings resulted in lattices that describe (partial) orders of classifiers according to (non-) overlapping parts of the input signal they process.
These results coincide with several prior publications on understanding the operations performed by deep neural network.
Zeiler \etal~\cite{zeiler2014visualizing} discovered that AlexNet is highly sensitive to local structures.
This can be understood in part as a consequence of the reduced input signal that AlexNet uses.
Raghu \etal~\cite{raghu2016expressive} showed that parameters on shallower layers have a higher impact on the final prediction than deeper layers.
Shallower layers are closer and hence more exposed to the entire signal and can therefore drop more of it.
Modifying shallow layers alters the amount and the kind of signal that goes into the network, affecting prediction scores significantly.
The work of Montavon \etal~\cite{montavon2017explaining} shows relevance reconstruction maps that are coarser for CaffeNet than for GoogleNet.
This phenomenon is consistent with other work~\cite{lapuschkin2016analysing} and can now be understood from the point of view of the input signal.
In one of their latest experiments, Bau \etal~\cite{netdissect2017} trained a variant of AlexNet with wider layers and global average pooling to explore its interpretability.
Despite all those changes, the accuracy was similar to the original architecture even after increasing the number of filters of the last convolutional layer by a factor of 4 and 8.
They suggest that the capacity of the network has been exhausted although, by definition, more filters indicate a higher capacity.
We propose a complementary idea: adding more filters in deeper layers do not affect the performance not because the capacity of the model is exhausted, but because the input signal is.
In other words, deeper layers already interpret all the signal that is available to them.

\section{Discussion \& Future Work}
\label{sec:discussion}

In this work, we introduced a novel, alternative way to understand the behavior of deep neural networks by studying the reconstruction performed by autoencoders that were fine-tuned to suit their needs.
This setup allowed us to analyze the amount of signal that a network uses before it enters the model itself.
Our approach is fundamentally different from previous work since we did not focus on measuring the behavior of intermediate or end results (\eg, through bias-variance metrics or activation maximization analysis).

We propose a training scheme for the autoencoder that ensures excellent generalization and reproducible results by using the YFCC100m as dataset for pre-training.
Furthermore, we use the resulting model as basis for further fine-tuning of the decoder with gradients from different pre-trained image classifiers.
By looking at the response of classifiers when different auto-encoded images are fed, we were able to establish a relation of order between classifiers that depends on the input signal.
We presented evidence of this underlying pattern by using formal concept analysis and validate our findings by measuring the information contained in the different image reconstructions.

\vspace{-1.5em}
\paragraph{Additional Findings:}
There are some further observations that spawn from our proposed method that we like to highlight.

The two-stage training strategy for autoencoders has influenced these networks to learn denoising operations.
Said function is effective because it favors the preservation of parts of the input signal that are used by the classifier, increasing the overall tolerance to noise within individual samples.

The amount of signal used by most classifiers is small compared to the amount of signal that is available from the input.
High performing image classifiers like ResNet can use as little or less than 10\% of the original input.
This can be seen as a beneficial property, as less evidence is required to make a correct prediction.
Possible downside of this reliance on little evidence is that small changes to relevant parts of the input, also known as adversarial examples, can change the prediction.
Classifiers that take advantage of redundant information are more robust to changes that were unaccounted for during training.
Note also that the amount of input signal does not correlate with the number of parameters or performance.

However, the relevant portion of the input signal does follow a general, distributed pattern as seen by the global checkerboard artifacts that appear in reconstructions produced by some AEs. In other words, these patterns to not depend on the specific content of each image, but rather to general patterns (\eg darken bright areas, increased distance between colors).

\vspace{-1.5em}
\paragraph{Future Work:}
As next steps, we want to compare the denoising properties of fine-tuned autoencoders with other architectures and training strategies that are explicitly designed to denoise.
Moreover, we would like to explore alternatives to SegNet as autoencoder that are more efficient and light-weight.
Finally, we are not yet able to point to specific points in the architecture of a deep network that are responsible for losing signal.
A better understanding of how the flow of signal through a network can be controlled will allow for a more principled approach to design future architectures.

\paragraph{Acknowledgments: }This work was supported by the BMBF project DeFuseNN (Grant 01IW17002) and the NVIDIA AI Lab (NVAIL) program.
We thank all members of the Deep Learning Competence Center at the DFKI for their comments and support.

{\small
\bibliographystyle{ieee}
\bibliography{s2snet}
}

\end{document}